\documentclass[10pt, a4paper]{article}

\usepackage{lrec-coling2024} 
\usepackage{tikz}
\usetikzlibrary{positioning, arrows.meta}
\usepackage{graphicx}
\usepackage{float} 
\usepackage{subcaption}

\title{Aspect-based Sentiment Evaluation of Chess Moves (ASSESS): an NLP-based Method for Evaluating Chess Strategies from Textbooks}

\name{Haifa Alrdahi and Riza Batista-Navarro
}

\address{Department of Computer Science, The University of Manchester, UK \\
haifa.alrdahi@manchester.ac.uk, haifa.alrdahi@gmail.com, riza.batista@manchester.ac.uk
}

\abstract{
The chess domain is well-suited for creating an artificial intelligence (AI) system that mimics real-world challenges, including decision-making. Throughout the years, minimal attention has been paid to investigating insights derived from unstructured chess data sources. In this study, we examine the complicated relationships between multiple referenced moves in a chess-teaching textbook, and propose a novel method designed to encapsulate chess knowledge derived from move-action phrases. This study investigates the feasibility of using a modified sentiment analysis method as a means for evaluating chess moves based on text. Our proposed Aspect-Based Sentiment Analysis (ABSA) method represents an advancement in evaluating the sentiment associated with referenced chess moves. By extracting insights from move-action phrases, our approach aims to provide a more fine-grained and contextually aware `chess move'-based sentiment classification. Through empirical experiments and analysis, we evaluate the performance of our fine-tuned ABSA model, presenting results that confirm the efficiency of our approach in advancing aspect-based sentiment classification within the chess domain. This research contributes to the area of game-playing by machines and shows the practical applicability of leveraging NLP techniques to understand the context of strategic games.
 \\ \newline  }

\begin{document}

\maketitleabstract

\section{Introduction}


Over the years, chess has long served as a testbed for evaluating the performance of various algorithms \cite{toshniwal2022chess}. One example is AlphaZero, which demonstrates the use of chess as a benchmark to evaluate the algorithmic performance of a self-learning algorithm \cite{silver2018general}. 
Most notably in recent years, artificial intelligence (AI) has been leveraged in the chess domain. For example, sentiment analysis has been used to evaluate chess moves based on commentaries \cite{kamlish2019sentimate}. Another example is predicting the next chess move based on patterns learnt from game databases, i.e., structured data \cite{noever2020chess}. These efforts highlight the benefits of applying AI to analyse moves in chess, a domain that has often been used to simulate real-world decision-making.

However, chess knowledge and strategies explained in free text has been under-explored, where limited research has investigated the ability and usefulness of evaluating chess moves expressed in unstructured data. Recently, a novel dataset, LEAP, was introduced \cite{alrdahi2023learning}.
It was derived from chess textbooks and includes structured (chess move notations and board states) and unstructured data (textual descriptions), aimed at teaching models about chess strategies by analysing descriptions of grandmaster games using sentence-level sentiment analysis. Drawing inspiration from the LEAP dataset, this work investigates the performance of a modified aspect-based sentiment classification method on new datasets---annotated at a finer-grained level---that were created to evaluate chess moves referenced in free text. We show that embracing the semantics of the chess domain to evaluate a move (i.e., the aspect) expressed in move-action phrases is a promising approach. This finer-grained analysis provides more detailed insight into the opinions expressed about the moves, contributing to a more comprehensive understanding of sentiments within the chess context, especially in multiple-aspect scenarios. We compared the performance of the proposed approach with an Aspect-Based Sentiment Analysis (ABSA) baseline approach and show that adopting a context-rich model with move-action phrase representations improves the results of sentiment analysis models. The contributions of this study are as follows:
\begin{itemize}
\item Creating a dataset from a chess-teaching textbook with fine-grained annotations. These include annotations of text spans pertaining to chess moves, players and predicates (verbs). The verbs are used to describe the actions and strategies involved in playing chess moves. Importantly, the sentiment expressed towards a given move is also annotated. 
\item Modifying the standard definition of `aspect' in a traditional ABSA approach by considering a player-predicate-move triple as an aspect. Adopting this definition, we designed a new ABSA method as a function for evaluating moves, which is the first attempt at exploring such an approach in the chess domain.
\item Training a RoBERTa model \cite{liu2019roberta} on the new task and evaluating its ability to choose strategic moves by using the ABSA evaluation function. Our modified method boosts model performance results on our datasets.
\item Using empirical evaluation to measure the reliability of our proposed method against Stockfish, a powerful search-based chess engine and tool for evaluating chess moves.
\end{itemize}
\section{Related Work}
Learning algorithms for playing chess have thus far overlooked the potential of obtaining knowledge from chess-teaching textbooks. Instead, knowledge is typically obtained from databases of chess moves, such as in the case of DeepChess \cite{david2016deepchess}. Such approaches are reliant on large, curated structured datasets capturing the knowledge of experts \cite{Schaigorodsky2016}, the production of which is often laborious and time-consuming 
and lacks explainability of the decision-making process in relation to the move.  AlphaZero, the board-based reinforcement learning algorithm, achieved remarkable results in both Chess and Shogi without relying on extensive domain knowledge beyond the fundamental rules of the games \cite{silver2018general}. However, this type of algorithm lacks human intuition or understanding of the game beyond what it learns through self-play. This can sometimes lead to unconventional strategies or a lack of understanding of traditionally accepted strategies in these games. Additionally, the resources required for the kind of intensive training that AlphaZero undergoes are computationally demanding and expensive, which limits its accessibility \cite{nechepurenko2020comparing}. 

Recent advances in natural language processing (NLP) such as the development of contextual embeddings and transformer architectures have boosted the performance of NLP-based models in many domains and tasks \cite{vaswani2017attention, devlin-etal-2019-bert}. This has provided opportunities to explore approaches that deviate from traditional ones that rely on chess engines,  which require extensive game state analysis to evaluate moves. 
For instance, a large language model (LLM) was trained on 10 million games annotated with action-value pairs from the Stockfish engine \cite{ruoss2024grandmaster}. The effectiveness of the LLM was evaluated on the basis of its ability to select the move with the highest value for any given position. The authors claimed that this model has the potential of achieving grandmaster-level chess-playing, without relying on explicit search algorithms, by predicting action-values directly from board states. However, this approach still lacks explainability as it does not require the LLM to provide any context. Instead, the model was aimed at mimicking the game-playing capabilities of Stockfish, which relies on search algorithms.

Nevertheless, various studies have shown that using context expressed in natural language as an alternative approach to overcome the above-mentioned limitations has improved the performance of AI systems. Previous work focussed mainly on extracting actions from sentences with short, direct instructions using a model with long short-term memory recurrent neural networks (LSTM-RNNs) \cite{mei2016listen}. However, such an approach does not have the ability to evaluate the outcome of the action, as it is applied to direct instructions only. 

Traditional sentiment analysis has been one of the few first attempts at applying NLP techniques to evaluate chess moves. SentiMate \cite{kamlish2019sentimate} presented an approach to chess move evaluation using NLP, where the model employs classifiers to determine move quality from commentary datasets, and a convolutional neural network (CNN)-based sentiment analysis model trained on chess commentaries. Although this method suggests the potential of NLP in improving decision-making processes as part of game strategies, it, however, offers limited insights when applied to extensive texts such those in chess textbooks. Chess commentaries typically focus on specific moves in given board states, while textbooks cover a range of moves. Standard sentence-level sentiment analysis methods are limited in that they capture sentiments towards individual moves only, without considering possible subsequent moves, thus limiting their capability as an evaluation function. 

Recent research has shifted to fine-tuning generative LLMs on extensive chess game data for move recommendation, such as the study by \citet{noever2020chess}. Their work demonstrates that GPT-2 can learn complex game play, where the model successfully learnt to generate plausible chess moves and strategies without direct intervention or heuristic guidance. Nevertheless, while the proposed approach achieved over 90\% accuracy in generating legal chess moves and reproducing classic chess openings, it neglects detailed knowledge of the textual content and lacks explanatory depth for specific move choices. Furthermore, the use of the GPT-2 model as an alternative method for search-based algorithms has not yet answered the question of whether the model genuinely comprehends the game or merely replicates move patterns effectively. The more recent study in by \citet{deleo2022learning} offers a fresh perspective on the use of language models to understand and learn complex strategic games such as chess. The study focussed on the ability of the BERT model to interpret chess positions of the board states and moves encoded in the Forsyth–Edwards Notation (FEN) format, analysing its effectiveness in playing against advanced chess engines like Stockfish. 
The BERT model exhibited its capability to maintain game-play against Stockfish for an extended number of moves and demonstrated substantial move accuracy, especially in the game's opening stages. However, the same limitation observed in other previous work remains, in that the model's proficiency in understanding deeper strategic elements of chess that go beyond the generation of valid moves, remains uncertain.


\section{ABSA for Chess Move Evaluation}
This section describes the motivation for using an Aspect-based Sentiment Analysis (ABSA) approach in evaluating chess moves described in free text. 
ABSA is considered to be a fundamental NLP task due to it being able to deliver more targeted sentiment insights, critical for a range of applications, including market analysis and social media monitoring \cite{pontiki2016semeval, zhang2022survey}. 
Recognised as a means for Information Extraction (IE) applied to various decision-making scenarios \cite{pontiki-etal-2014-semeval, phan2021approach, sun2020applications, reynard2019harnessing}, this approach lends itself well to the analysis of strategic chess moves in a teaching context, particularly to determine the suitability of a move in the context of a given board state. Here, ABSA evaluates the moves, labelling them as `go-to', `avoid', or `neutral'.

Meanwhile, the LEAP corpus highlighted the challenge of handling multiple aspects within a single sentence \cite{alrdahi2023learning}. 
Investigating the description of moves within this corpus reveals a strong similarity in sentence structures, but understanding the different meanings requires careful interpretation. 
We argue that, for this purpose, knowledge infusion (i.e., the incorporation of structured knowledge) can help models achieve a deeper understanding of the domain, leading to more accurate and context-aware predictions \cite{sheth2019shades}.
Hence, we consider a chess move as a distinct aspect, whereby we integrate the verb (predicate) indicating whether the move is `go-to' or `avoid'. Within the context of chess, each move is associated with a player performing the action. Therefore, we also incorporated the player into the aspect, resulting in a player-predicate-move triple. We hypothesise that this aspect, which we refer to as the `move-action phrase', can assist a classification model in differentiating between different sentiments associated with multiple aspects in the same sentence.

To validate the above hypothesis, we performed an ablation study that compared models, both with and without this novel aspect formulation. Additionally, every move is played based on a specific type of aim or action, such as attacking or protecting a piece. These actions are expressed in different predicates by grandmaster players, where different predicates could mean the same type of action. We design unsupervised clusters of action types and infused the original sentences with the action type as additional semantic information. By adding this contextual information, we are enriching the model's input with external structured knowledge (the action type), which is not inherently part of the original sentence data. In this paper, we are studying the effect of knowledge infusion that in theory should enhance a model's understanding by providing additional context. 

\section{Dataset Description}
\subsection{Annotation Process}
Given that the chess domain has well-defined terminologies to express moves, we developed a rule-based named entity recognition (NER) method based on regular expressions (regexes) to extract mentions of pieces, players, moves and move sequences, where the lattermost are expressed in the Standard Algebraic Notation (SAN) chess notation format. Upon manual review, we noticed that each selected move represents a distinct type of action, aimed at achieving a particular purpose or strategy, which is expressed in natural language through the use of verbs. We then crafted an annotation scheme at the move-action phrase level (where a move can act as either the subject or object of a verb) for aspect-based sentiment classification. We took sentences in the LEAP corpus that were determined as topic-relevant, i.e., pertaining to strategic moves \cite{alrdahi2023learning}, and applied the WordNet English lexicon \cite{miller-1994-wordnet} to identify verbs within sentences. The annotation of all move-action phrases was conducted by duplicating each sentence according to the number of verbs identified within it. 

\begin{figure*}[h]
\centering
\includegraphics[width=0.5\textwidth,height=10cm]{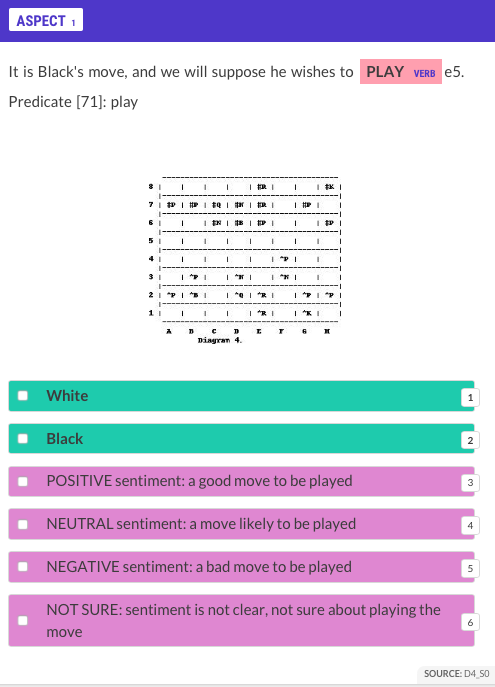}

\caption{The annotation interface as shown in Prodigy.}
\label{view}
\end{figure*}

As suggested in the literature, access to the state of the board during the evaluation of moves improves understanding of the context, leading to a more precise evaluation of the outcome of a move in the context of a given board state \cite{alrdahi2023learning, beinborn-etal-2018-multimodal, pezzelle-etal-2020-different}. Therefore, in annotating our data, we used Prodigy,\footnote{\url{https://prodi.gy/}} which allows for annotation based on multi-modal information. 
Each example in our dataset is presented to annotators in an interface that is divided into four sections, as shown in Figure~\ref{view}.
These sections are necessary in understanding the context of a chess game and ensuring accurate annotation:

\begin{enumerate}
    \item An image depicting the chess board's state, aiding in visualising the context for the move under consideration and facilitating visualisation of state changes based on sentences describing multiple moves.
    \item A text description of potential moves, with the relevant predicate emphasised in bold and tagged as `VERB' to direct the annotator's attention to the move's significance and its sentiment. For example, in the sentence \emph{``It is Black's move, and we will suppose he wishes to play e5''}, the predicate \emph{``play''} is highlighted.
    \item A list of options corresponding to player names, namely, ``White'' and ``Black'', which are for indicating the player making the move.
    \item Sentiment labels for indicating the sentiment toward the move: 
    \begin{description}
     \item[`Positive'] for advantageous moves.
     \item[`Negative'] for disadvantageous moves.
     \item[`Neutral'] for moves with neither a positive nor a negative effect.
     \item[`Not sure'] for when the sentiment towards a move is unclear.
    \end{description}
\end{enumerate}

The sentence shown in Figure~\ref{view}, when annotated, has Black as the player and `Neutral' as the sentiment for the aspect \emph{Black-play-e5}. 
Meanwhile, multiple aspects can be found in the example sentence \emph{``Before bringing the discussion of the Queen's Pawn opening to a close, I may remark that in tournaments it has become usual for White not to play c4 at once, but to play Nf3 as a preliminary, in order to avoid the complications of the Queen's counter gambit.''}
In this case, the aspect \emph{White-play-c4}, is labelled as `Negative', and the second aspect \emph{White-play-Nf3} is labelled as `Positive'. 

To closely replicate the chess environment, annotators were instructed to: (1) select the player making a move, (2) identify the move, and (3) determine the sentiment towards the player-predicate-move triple. The dataset, consisting of 726 sentences, was annotated by the lead author of this paper and one additional annotator with expertise in NLP and sentiment analysis, and familiarity with chess terminology. 
To enable measurement of inter-annotator agreement (IAA), 20\% of the total number of sentences was set aside as a common subset that was annotated by both annotators, albeit independently.
Annotator agreement was measured using Cohen's Kappa metric and was determined to be 65\% (substantial agreement).
Each annotator was then tasked with annotating a further (non-overlapping) subset with 10\% of the total number of sentences, which are unique to that annotator.
The annotations resulted in 437 `Positive', 153 `Negative' and 133 `Neutral' labels, with only three instances of uncertainty that were subsequently removed. The dataset was divided into training (70\%), validation set (10\%) and testing (20\%) sets.

\subsection{Data Augmentation}

Confronted with the known challenge of limited resources for manually creating training data \cite{zhang2022survey} and the need to address the LEAP dataset's imbalance, we employed an oversampling technique from the nlpaug library.\footnote{\url{https://github.com/makcedward/nlpaug/tree/master}} This approach is based on back-translation, leveraging two translation models to convert sentences from English to German, and then back to English. This method effectively generated additional synthetic data for model training, to enhance the diversity of the dataset without manually curating content \cite{wei-zou-2019-eda, feng2021survey}. The augmented sentences retained the original sentence meaning and sentiment, albeit with minor contextual variations. Unlike prompt-based generation, which might produce random outcomes or undesirable sentence alterations, this method offered controlled sentence generation. We applied this technique to achieve a balanced distribution of labels in the training set as shown in Table \ref{tab:label_counts} and manually checked the quality and correctness of the generated sentences. 
\begin{table}[!ht]
\begin{center}
\begin{tabularx}{\columnwidth}{|l|X|X|}

      \hline
      \textbf{Label} & \textbf{Original} & \textbf{Over-sampled} \\
      \hline
Positive & 288 & 288 \\
Negative & 117 & 234 \\
Neutral & 100 & 200 \\
\hline
\end{tabularx}
\caption{Distribution of training set labels.}
\label{tab:label_counts}
\end{center}
\end{table}

\section{Transformer Models for ABSA}
In this section, we describe the various transformer-based language models that we built upon in order to construct ABSA classifiers.

\subsection{Generic ABSA Models}
Firstly, we evaluated the vanilla RoBERTa-base (VRB) model on the aspect-based sentiment classification task, using two existing general-domain corpora: the Restaurant and Laptop dataset from SemEval 2014 \cite{pontiki-etal-2014-semeval}, and the MAMS dataset \cite{jiang-etal-2019-challenge} which bears similarities with LEAP in that a single sentence could bear multiple sentiments. To fine-tune VRB on the Restaurant and Laptop datasets, we used baseline parameters (seed of 42, batch size of 4, a learning rate of 3e-05, and no weight decay). Additionally, VRB performance on the MAMS dataset was evaluated under two conditions: using the baseline settings above and following the hyperparameters (batch size of 8) of CapsNet-BERT, a state-of-the-art ABSA model \cite{jiang-etal-2019-challenge}. Performance was measured by taking the mean of the micro-averaged F1-scores across five runs, selecting the best epoch score for each run. VRB demonstrated competitive micro-averaged F1-scores against leading models on SemEval 2014 Restaurant and Laptop datasets (85.68\% and 80.05\%, respectively), and achieved comparable results on the MAMS dataset (84.29\%). VRB demonstrated strong baseline performance without hyperparameter optimisation, additional features or additional training data. These scores encouraged us to adopt RoBERTa as our primary architecture for move evaluation experiments. 

\subsection{Fine-tuned ABSA Models}
Based on similar work and recommendations reported in the literature \cite{xu-etal-2019-bert, 9412167, rietzler-etal-2020-adapt}, fine-tuning a language model on domain-specific data significantly improves the model's understanding of relevant domain knowledge. This specialised training phase adjusts the model's understanding to learn the context and terminology within a specific domain, hence facilitating more precise and informed predictions or analyses. We fine-tuned the VRB model (henceforth referred to as the FT-RB model) using synthesised chess sentences in the LEAP dataset \cite{alrdahi2023learning} and chess commentaries collected by \citet{jhamtani-etal-2018-learning}.

We approached the task of move evaluation as a sequence classification problem, representing the dataset as $D = \{(X_i, Y_i)\}_{i=1}^{|D|}$, where $X_i$ is an input sentence and $Y_i$ is the corresponding true label for the $i^{th}$ instance. In our sequence classification task, both the sentence and a specific aspect are treated as part of the input $X$. The sentence $X$ is fed into the model encoder $Enc(X)$ to derive contextual features. The sentence's final hidden state is encapsulated by the special token $CLS(X)$ through a dense layer with a softmax function for predicting $Y$.

\begin{figure*}[!ht]
    \centering
   \begin{subfigure}[b]{\linewidth}
        \centering
        \includegraphics[width=0.65\linewidth]{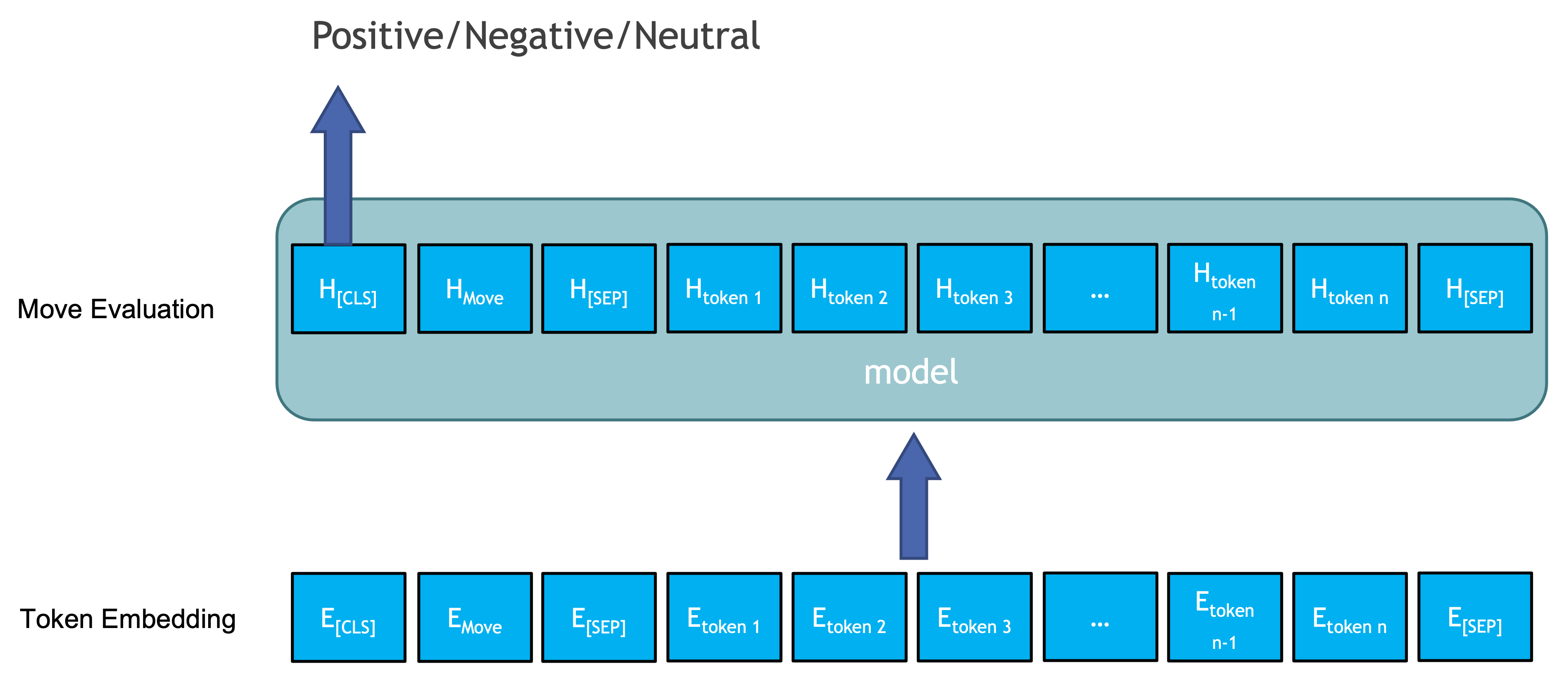}
        \caption{`move only' aspect}
        \label{fig:sub1}
    \end{subfigure}
    \rule{\linewidth}{0.4pt}
    
    \begin{subfigure}[b]{\linewidth}
        \centering
        \includegraphics[width=0.75\linewidth]{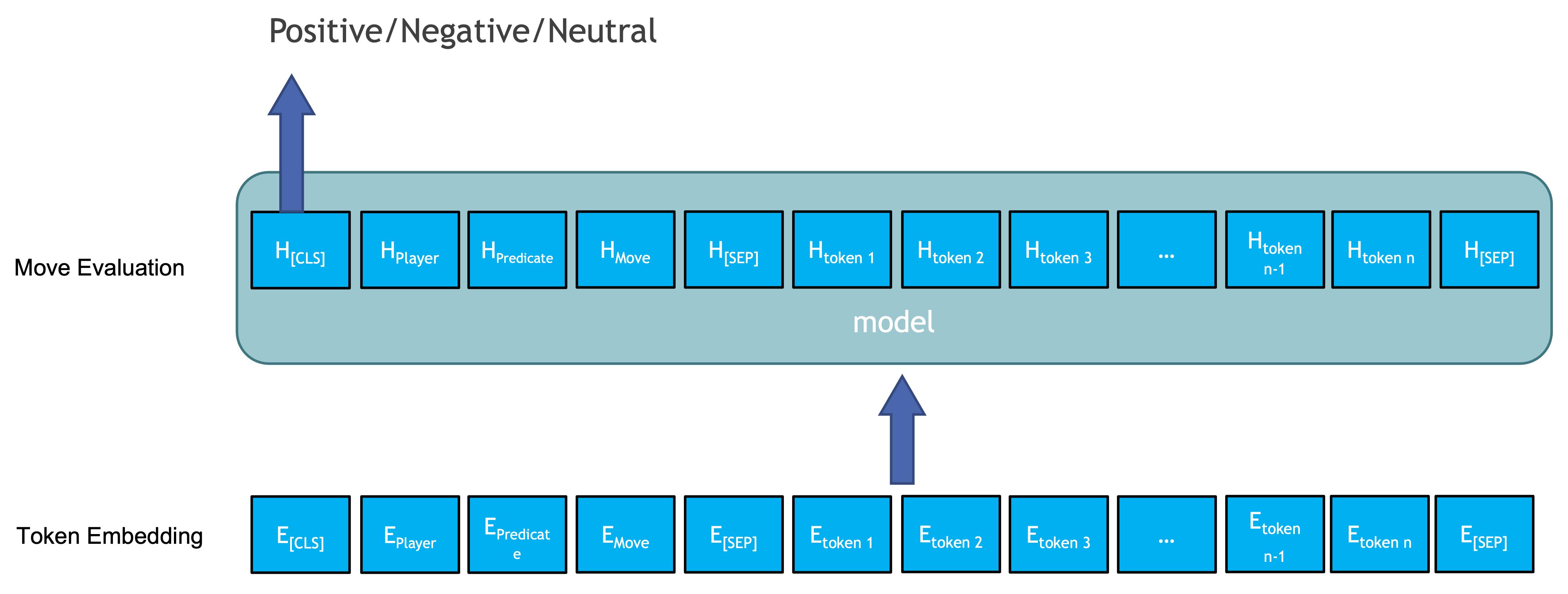}
        \caption{`move-action' aspect}
        \label{fig:sub2}
    \end{subfigure}
    \rule{\linewidth}{0.4pt}
    
    
    \begin{subfigure}[b]{\linewidth}
        \centering
        \includegraphics[width=0.85\linewidth]{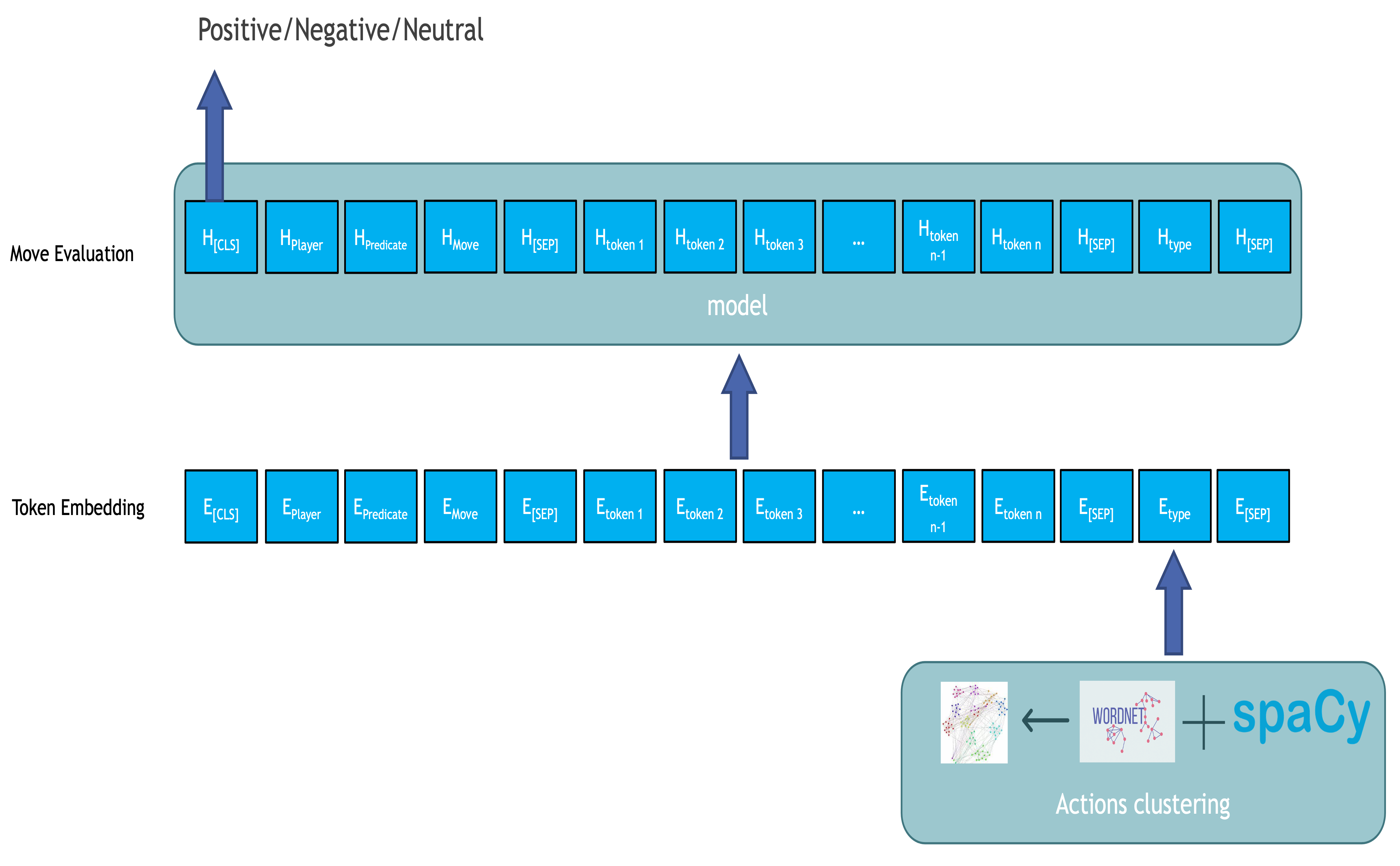}
        \caption{`move-action' aspect with semantic information}
        \label{fig:sub4}
    \end{subfigure}
    \caption{Different types of knowledge infusion for aspect-based sentiment classification  models.}
    \label{fig:sent_embed}
\end{figure*}

Three types of sentence embeddings were evaluated as representations of an infused input sentence, illustrated in Figure \ref{fig:sent_embed}. In Figure (a), the embedding $X$ represents `move only' as the aspect.
In Figure (b), $X$ represents the `move-action' phrase as the aspect.
Lastly, in Figure (c), $X$ corresponds to the embedding of the `move-action' phrase as the aspect, but enriches the sentence with additional information on the move-action type. All additional inputs in the sentence were separated with the special token \texttt{[SEP]}. We defined five types of move-actions: Attack, Capture, Defend, Protect, and General Move; we refer the reader to Table \ref{tab:example} for their definitions and corresponding examples. 
To automatically classify move-actions according to these types, we employed an unsupervised learning approach that leveraged semantic relationships from WordNet, including synonyms and definitions to group the verbs. The Chinese Whispers graph clustering algorithm \cite{biemann2006chinese} was applied with the aid of Gephi \cite{bastian2009gephi} to group verbs based on cosine similarity, using a seed value of 50 and running the clustering for 50 iterations. Through experimentation, we established a minimum similarity threshold of 40 for clustering data points.

\begin{table}[!ht]
\begin{center}
\begin{tabularx}{\columnwidth}{|l|X|X|}
\hline
\textbf{Move type} & \textbf{Definition} & \textbf{Example}\\
\hline
Attack & Playing a move to attack an opponent piece & \emph{White can attack Bishop with his Rook} \\
\hline
Capture & Playing a move to capture an opponent piece & \emph{White captures Bishop with his Rook} \\
\hline
Defend & Playing a move to defend a piece under an attack. & \emph{Black defends his Bishop by pushing it to d7} \\
\hline
Protect & Playing a move to protect a piece from future attack. & \emph{Black can protect his Bishop by pushing pawn to c6} \\
\hline
Move & A general move, (placing a piece from one position into another) without explicit intention of a purpose. & \emph{White plays 5.b3 before castling} \\
\hline
\end{tabularx}
\caption{\label{tab:example} Definitions of chess move-action types with examples.}
\end{center}
\end{table}

\section{Evaluation}
Our experiments were run five times, and the average F1-score was reported. In our ablation study, we tested different types of sentence embeddings by either removing or including the move-action type. We also examined how swapping the two types of aspects, i.e., `move only' and our modified `move-action' phrase,  affects the results. We used default hyperparameter values: seed = 42, evaluation and training batch size per device = 4, learning rate = 3e-05 and weight decay = 0.0.

\subsection{ABSA Results}
To evaluate the impact of our aspect formulation and the effects of knowledge infusion, we compare the vanilla RoBERTa-base (VRB) and fine-tuned RoBERTa-base (FT-RB) models on the task of aspect-based sentiment analysis as a means for chess move evaluation. We report the results on the original dataset in Table \ref{tab:results1} and on the over-sampled dataset in Table \ref{tab:results2}. 

The F1-scores achieved by the two models indicate that despite the inherent complexity of the task, the FT-RB model demonstrated an improvement in its F1-score. This improvement can be attributed to the modification of the model's weights, facilitated by the incorporation of domain-specific knowledge data. Furthermore, the modification of aspect definition we designed contributed towards enhancing the performance of the FT-RB model. This adjustment enabled the model to comprehend the context more effectively and concentrate on the `move' that is being evaluated, even where multiple aspects are being discussed in the sentence. 
However, despite the higher F1-scores observed when using the original dataset, it is worth noting that the model failed to accurately identify the minority classes, specifically the `Negative' and `Neutral' class labels, and the relatively high score is due to the larger number of examples labelled with the `Positive' class label. Using oversampled data, we observed an improvement in the model's capability to grasp and categorise these minority classes, gaining an increase of 20\% to 30\% in terms of F1-score for the `Negative' and `Neutral' classes. 
The results obtained by the FT-RB model trained on the oversampled data demonstrate the potential of the model to learn minority classes, hence we are considering it as our preliminary proposed model for this challenging task. 

Interestingly, adding the type of move-action to the input slightly decreased the performance of both models. This decrease can be attributed to the added complexity of the information in the input, which the models found difficult to interpret. Even though, intuitively, the move-action type adds further knowledge to allow humans to interpret the reason for playing the move, it did not aid the aspect-based sentiment classification models.


\begin{table*}[!ht]
\centering
\begin{tabular}{|c|c|c|}
\hline
\textbf{Embeddings} & \textbf{VRB} & \textbf{FT-RB} \\ \hline
move only & 54\% & 55\% \\ \hline
move-action phrase & 55\% & 62\% \\ \hline
move-action phrase with type & 55\% & 59\% \\ \hline
\end{tabular}
\caption{Averaged F1-scores obtained on the original dataset, using different types of input sentence representations.}
\label{tab:results1}
\end{table*}

\begin{table*}[!ht]
\centering
\begin{tabular}{|c|c|c|}
\hline
\textbf{Embeddings} & \textbf{VRB} & \textbf{FT-RB} \\ \hline
move only & 41\% &  50\% \\ \hline
move-action phrase & 50\% & 55\% \\ \hline
\end{tabular}
\caption{Averaged F1-scores obtained on the oversampled dataset, using different types of input sentence representations.}
\label{tab:results2}
\end{table*}

\subsection{ABSA vs. Stockfish Analysis}
Deviating from the usual game-level evaluation of chess agents, we assessed the extent to which our proposed ABSA model can evaluate a move based on a given board state.
This is because our corpus does not discuss a complete game from its start to end, but focusses only on specific strategic moves. A board state is represented in the Forsyth-Edwards Notation (FEN) format, which can capture the placement of pieces on the board, turn-to-move, castling availability, and other basic chess rules.  We extracted the FEN board and the chess moves discussed in the text, and integrated them into the Stockfish 16 engine to obtain the probability of the move leading to a win, a loss or a draw. Unlike other games where the outcome at the end is only a win or a loss, chess recognises the draw as a third possible outcome. Draws are common in high-level chess matches, where players often have similar strengths and capabilities, and neither side has achieved an advantage to claim a win.

In Stockfish, the engine skill was set to 8, the Elo rating to 2400 (grandmaster level) and the search depth to 10. These settings were selected to establish a baseline of how much we can rely on the evaluation described in the text, as if the text-based sentiment evaluation is equivalent to depth search. 
In cases where the aspect is a sequence of moves, we take Stockfish engine's evaluation of the first move in the sequence. We excluded counterfactual statements, moves that were incorrect as a result of issues encountered during optical character recognition (OCR) when the textbook was digitised, and implicit moves. An example of an implicit move is the triple \emph{Black-play-King away from his file} from the sentence \emph{``White has no time to double Rooks, because if he does so, after his Re2 Black would play the King away from his file and allow the Knight to escape''.} 

\begin{figure}[h]
\centering
\includegraphics[width=9cm, height=6cm]{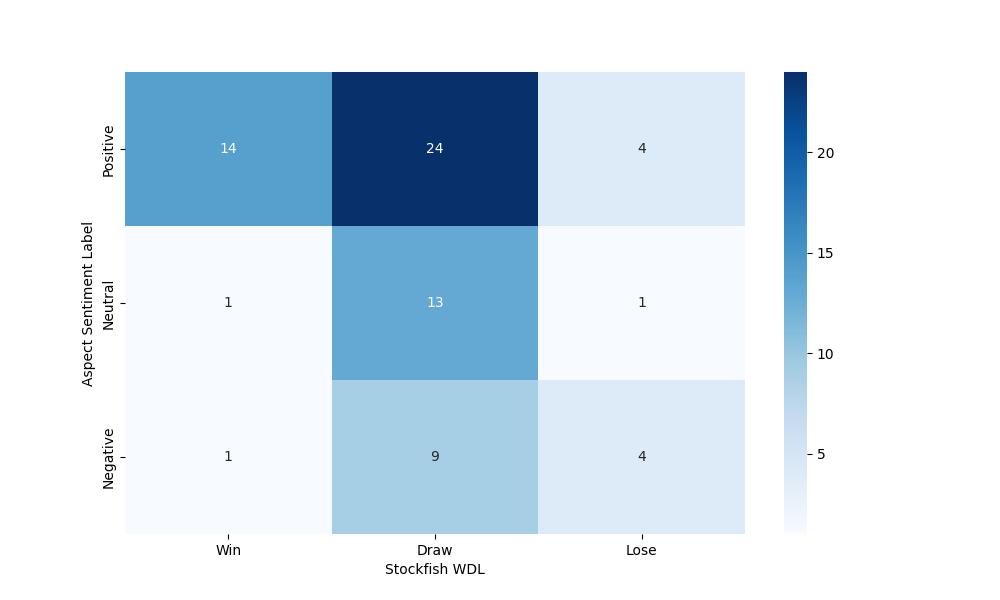}
\centering
\caption{Heatmap based on a sample (10\%) to visualise the correlation between the sentiment labels predicted by our baseline ABSA model for a move (`Positive', `Neutral', `Negative'), and the probable outcome (`Win', `Draw', `Lose') of the move provided by the Stockfish engine.}
\label{Fig:heatmap}
\end{figure}

The heatmap in Figure \ref{Fig:heatmap} presents the number of times each sentiment label has the highest score in one of the categories: `Win', `Draw', `Lose'. The `Positive' and `Negative labels have a significantly higher association with the `Draw' category than with the `Win' or `Lose' categories, respectively. The relatively low count for the `Lose' category being associated with the `Positive' and `Neutral' labels indicates that the sentiment labels are relatively aligning with the outcome of the search-based algorithm, even though the ABSA model did not have access to the board state during sentiment classification. 

It can be seen that a substantial number of moves labelled as `Neutral' are correlated with the `Draw' category, suggesting that these moves maintain the balance of the game where neither side has an advantage. The distribution of the `Negative' label is balanced across all Stockfish outcomes, with a slight preference for `Draw' and `Lose'. This implies that negative sentiments are indicative of more challenging or risky positions, which might lead to either a loss or a stabilising effort towards a draw. Finally, a higher number of positive sentiment labels are associated with winning outcomes and are less frequently associated with losses. Overall, the ability to correlate sentiment labels with the game outcomes suggests that the sentiment expressed in the description of chess moves can be predictive of the move's effectiveness without requiring deep analysis by a chess engine. This analysis supports our hypothesis that ABSA can be an evaluation function for chess moves described in text, offering a novel approach to understanding and predicting the implications of chess strategies through aspect-based sentiment classification.

\subsection{Error Analysis}
We identified multiple factors that might have affected the performance of the ABSA model and might have led to disagreements between its predictions and the outcomes provided by Stockfish. Firstly, some errors were made by our automatic rule-based move extraction method, leading to the linking of predicates to incorrect moves, which in turn, resulted in wrongly predicted sentiment labels. A key limitation of our current work is that aspect extraction is not learnt jointly with the sentiment classification task, whereas end-to-end systems (i.e., models trained simultaneously on aspect extraction and sentiment classification) have reportedly obtained better performance. 

An additional limitation is the model's under-performance in capturing the strategic depth of chess moves not explicitly described in SAN nor natural language, such as implicit threats or long-term strategies not explicitly mentioned in the text. An example sentence that contains such implicit information is: \emph{``In Diagram 13, White derives no advantage from being the exchange to the good, for the Rook has no file which could be used to break into the Black camp.''} Even with access to the board state diagram and despite the sentiment being clearly negative, it is difficult to determine what \emph{``exchange''} refers to. The model could be improved to become a powerful chess agent by using a reinforcement learning approach to take advantage of feedback, which we leave for future work. The feedback could involve integrating a chess engine's evaluations alongside features from natural language to provide a more comprehensive understanding of the positions. This would enable the model to consider both explicit descriptions and the underlying strategic implications of moves, offering a fuller analysis of chess strategies from textual descriptions.


\section{Conclusion}
This study is part of ongoing research that investigates approaches to evaluating chess moves described in textbooks. We have demonstrated, based on a small-scale dataset, the potential of using text-based resources to evaluate strategic chess moves. We introduced a novel method for evaluating chess strategies using NLP, specifically focussing on aspect-based sentiment classification of chess moves described in textbooks. 
This involves creating a new annotated dataset drawn from the chess literature, modifying the definition of aspect (in ABSA) to include both player and move-action phrases, and training a RoBERTa-base sentiment classification model for strategic move evaluation.

Our research presents baseline results for this new task, which demonstrate the potential of NLP to improve understanding and analysis of chess strategies. 
We believe that if enough text-based datasets are available, the model can potentially evaluate moves and engage in game-play. 

Finally, many studies have shown that grounding natural language with the environment results in more accurate decision-making \cite{kameko2015learning,matuszek2018grounded,alomari2017natural,karamcheti2017tale,branavan2012learning,luketina2019survey,he2016deep}. Hence, aligning the sentiment analysis with the environment---in this case, the chess board state---could facilitate more effective decision-making. As part of future work, our aim is to further explore the incorporation of the board state into the input for aspect-based sentiment classification. In this scenario, a move will be evaluated not only on the basis of the text but also while considering the current board state and the additional semantic knowledge represented by the action type \cite{zhang2022survey}. In addition, an ABSA model could enhance the game-playing experience by offering an explanation of the choice of moves through search-based chess agents, which we aim to explore next. 

\nocite{*}

\section{Bibliographical References}\label{sec:reference}

\bibliographystyle{lrec-coling2024-natbib}
\bibliography{lrec-coling2024-example}





\end{document}